\documentclass[nohyperref]{article}

\usepackage{microtype}
\usepackage{graphicx}
\usepackage{subfigure}
\usepackage{booktabs} 

\usepackage{hyperref}
\usepackage{multirow}

\usepackage[accepted]{icml2022}

\usepackage{amsmath}
\usepackage{amssymb}
\usepackage{mathtools}
\usepackage{amsthm}

\usepackage[capitalize,noabbrev]{cleveref}

\theoremstyle{plain}

\theoremstyle{definition}

\theoremstyle{remark}

\usepackage[textsize=tiny]{todonotes}

\icmltitlerunning{Topology-Aware Network Pruning using Multi-stage Graph Embedding and Reinforcement Learning}

\begin{document}

\twocolumn[
\icmltitle{Topology-Aware Network Pruning using Multi-stage Graph Embedding and Reinforcement Learning}



\icmlsetsymbol{equal}{*}

\begin{icmlauthorlist}
\icmlauthor{Sixing Yu}{isu}
\icmlauthor{Arya Mazaheri}{tud}
\icmlauthor{Ali Jannesari}{isu}

\end{icmlauthorlist}

\icmlaffiliation{isu}{Department of Computer Science, Iowa State University, Iowa, US}
\icmlaffiliation{tud}{Department of Computer Science, Technical University of Darmstadt, Darmstadt, Germany}

\icmlcorrespondingauthor{Sixing Yu}{yusx@iastate.edu}
\icmlcorrespondingauthor{Arya Mazaheri}{arya.mazaheri@tu-darmstadt.de}
\icmlcorrespondingauthor{Ali Jannesari}{jannesar@iastate.edu}

\icmlkeywords{Machine Learning, ICML}

\vskip 0.3in
]



\printAffiliationsAndNotice{}  

\begin{abstract}
Model compression is an essential technique for deploying deep neural networks (DNNs) on power and memory-constrained resources. However,  existing model-compression methods often rely on human expertise and focus on parameters' local importance, ignoring the rich topology information within DNNs. 
In this paper, we propose a novel multi-stage graph embedding technique based on graph neural networks (GNNs) to identify DNN topologies and use reinforcement learning (RL) to find a suitable compression policy. We performed resource-constrained (i.e., FLOPs) channel pruning and compared our approach with state-of-the-art model compression methods.
We evaluated our method on various models from typical to mobile-friendly networks, such as ResNet family, VGG-16, MobileNet-v1/v2, and ShuffleNet. 
Results show that our method can achieve higher compression ratios with a minimal fine-tuning cost yet yields outstanding and competitive performance.
The code is open-sourced at \url{https://github.com/yusx-swapp/GNN-RL-Model-Compression}.

\end{abstract}
\section{Introduction}
The demand for deploying DNN models on edge devices (e.g., mobile phones, robots, and self-driving cars) is expanding rapidly. However, the increasing memory and computing power requirements of DNNs make their deployment on edge devices a grand challenge. Thus, various custom-made DNN models have been introduced by experts to accommodate a DNN model with reasonably high accuracy on mobile devices~\cite{howard2019searching,tan2019efficientnet,zhang2018shufflenet,ma2018shufflenetv2,mehta2020dicenet,huang2018condensenet}. 
In addition to mobile-friendly deep networks, 
model compression methods such as network pruning, have been considerably useful by introducing sparsity or eliminating channels or filters. Nevertheless, it requires extensive knowledge and effort to find the perfect balance between accuracy and model size.

The main challenge of network pruning is to find the best pruning schedule or strategy. Furthermore, a pruning strategy for a given DNN is not transferable to a different DNN, demanding a customized per-network pruning strategy.
Recently, various pruning methods~\cite{he2018amc,yu2020agmc} have been proposed to automatically compress DNNs. However, they either use manually defined rules/embeddings, ignoring rich topological information, or do not consider topology changes while model compression.
Moreover, since RL-based methods~\cite{liu2020AutoCompress,he2018amc,yu2020agmc} usually use the pruned model accuracy as RL agent's reward function, a negative correlation emerges between the compression ratio and reward. Consequently, without any constraint, the RL agent tends to search for a tiny compression ratio to get a better reward. To address this problem and get the desired model size reduction, existing RL-based methods often need additional heuristic algorithms to adjust the pruning ratio. 

The computational representation of DNNs often contains various patterns (a.k.a. motifs) repeated throughout the network topology. For instance, MobileNetV2 involves 17 blocks, each following a similar graph and operation structure. 
The topology of such blocks can represent their states, allowing us to exploit their redundancy level and importance.
Such structural information inspired us to model a given DNN as hierarchical computational graphs and propose multi-stage graph neural networks (m-GNN) for DNN embedding. 
Additionally, we equipped m-GNN with a reinforcement learning agent (GNN-RL) to automatically search for the compression policy (i.e., pruning ratios). To avoid tiny compression ratios due to the negative correlation between the compression ratio and the RL agent's reward, we created a DNN-Graph environment for the GNN-RL agent. Such an environment allows the agent to continuously compress the DNNs until it satisfies the model size constraint. 
For each step of the compression, the DNN-Graph environment converts the compressed DNN to a graph. The graph is the environment state input to the GNN-RL agent. Once the compressed DNN satisfies the desired model size, the DNN-Graph ends the search episodes and uses the pruned DNN's accuracy as a reward for the agent.
We successfully performed FLOPs-constraint network pruning on various DNNs and achieved competitive results with the state-of-the-arts pruning methods.
More importantly, the experiments showed that the learned topology of a given DNN could be transferred to another DNN. Such a feature proves that graph embedding is applicable to network pruning.

In essence, this paper makes the following contributions:
\begin{itemize}
    \item A novel method for modeling DNNs as hierarchical graphs to exploit their topological and structural information for topology-aware network pruning.
    \item An efficient multi-stage GNN~(m-GNN) to learn hierarchical and transferable graph embeddings.
    \item A simpler yet efficient RL method based on the PPO algorithm for adaptive network pruning.
    \item Competitive results with the state-of-the-art model compression methods on various DNN models.
\end{itemize}

\section{Related Work}
Various studies focus on model compression and efficient deployment of DNNs, such as network pruning~\cite{han2015deep,he2018amc,li2020eagleeye,chin2020legr,guo2020dmcp,ye2020goodsubnet,zhuang2020neuron,tang2020scop,ning2020dsa,chen2021oto,lai2021parp,gao_network_2021,liu_learnable_2021,Wang2021cov,chen2020storage}, knowledge distillation~\cite{hinton2015distilling}, and network quantization~\cite{gholami2021survey_quantization}. Within the scope of this paper, we mainly focus on structured network pruning~\cite{Anwar2017Structured}, as it is not bound to special AI accelerators~\cite{zhang2018unstructured,Guo2016unstructured}. 
Uniform, shallow, deep empirical structured pruning policies~\cite{he2017handcraft_channel,Li2016handcraft}, the hand-crafted structured pruning methods, such as SPP~\cite{wang2017SPP}, FP~\cite{Li2016handcraft}, and RNP~\cite{Lin2017RNP} fall into the structured pruning category. 
However, such pruning policies often fail to work properly on new models and might lead to sub-optimal performance.
Recently, AutoML pruning algorithms~\cite{li2020eagleeye,chin2020legr,ye2020goodsubnet,chen2020storage,li2020dhp,chin2020legr,lin2020hrank,li2020group} offered better results with higher versatility, particularly the RL-based methods~\cite{liu2020AutoCompress,he2018amc,yu2020agmc}. Liu et al.~\cite{liu2020AutoCompress} proposed an ADMM-based~\cite{Boyd2011ADMM} structured weight pruning method and an innovative additional purification step for further weight reduction. He et al.~\cite{he2018amc} proposed AMC and used RL to predict each hidden layer's compression policy. However, they manually defined DNN embeddings, such as the number of input/output channels, parameter size, and FLOPs for the RL environment state vectors, and ignored the neural network's essential structural information. Yu et al.~\cite{yu2020agmc} modeled DNNs as graphs and introduced a GNN-based graph encoder-decoder to embed DNNs' hidden layers. Nevertheless, they performed layer-wise pruning based on simple layer embeddings, ignoring the global topology changes when pruning. Moreover, they construct simplified computational graphs for DNNs and do not take advantage of motifs in DNNs.

\textbf{Graph Neural Networks (GNN).}
GNN and its variants~\cite{kipf2017gcn,Schlichtkrull2018rgcn} can learn the graph embeddings and have been successfully used for link prediction~\cite{Nowell2007linkprediction} and node classification. However, these methods are mainly focused on node embedding and are inherently flat, which is inefficient to deal with the hierarchical data. In this paper, we aim to learn the global topology information from DNNs. Thus, we proposed multi-stage GNN (m-GNN), which takes advantage of the repetitive motifs available in DNNs. m-GNN considers the edge features and has a novel learning-based pooling strategy to learn the global graph embedding.

\textbf{Graph-based Neural Architecture Search (NAS).} Although this paper is not directly related to NAS, it is an active area of research wherein the computationally expensive operations are replaced with a more efficient alternative~\cite{li2021gnasr,li2021npas,Yao2021JointDetNASUY,wang2020apq}. 
Particularly, graph-based NAS methods apply GNN and use graph-based neural architecture encoding schemes to exploit neural network's topology. They model neural architecture's search spaces as graphs and aim to search for the best performing neural network structure~\cite{Guo2019NAS_NAT,Han2020NAS_oneshot,Dudziak2021BPR_NAS,chatzianastasis2021graph,ning2020generic}.
Such methods inspired us to exploit compression policy from the topology information of DNNs.

\section{Approach}
To prune a given DNN, the user provides the model size constraint~(i.e., FLOPs constraint). Although we perform FLOPs-constraint filter pruning, our method is not limited to FLOPs-constraint and can be easily extended to latency, MACs, or sparsity constraint compression.

\begin{figure}[t]
\begin{center}
\includegraphics[width=1\columnwidth]{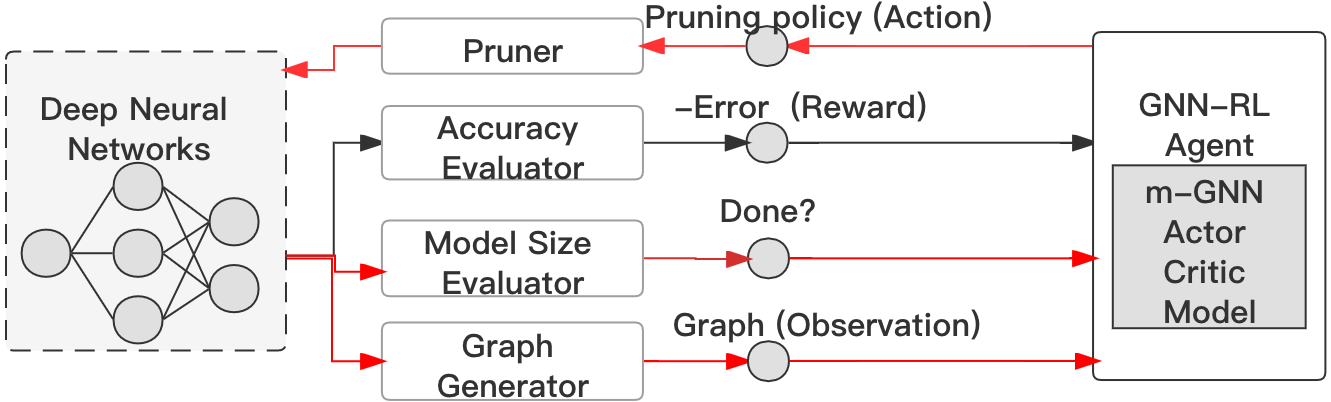}
\vskip -0.1in
\caption{An overview of DNN-Graph environment search episode.}
\label{fig:2}
\end{center}
\vskip -0.3in
\end{figure}

Figure~\ref{fig:2} illustrates the DNN-Graph search episode, which is essentially a model compression iteration. Red arrows show that the process starts from the original DNN. The model size evaluator first evaluates the size of the DNN. If the size is not satisfied, the graph generator converts the DNN into a hierarchical computational graph. Then, the GNN-RL agent leverages m-GNN to learn pruning ratios from the graph. The pruner prunes the DNN with the pruning ratios and begins the next iteration from the compressed DNN. 
Each step of the compression will change the network topology. Thus, the DNN-Graph environment reconstructs a new hierarchical computational graph for the GNN-RL agent corresponding to the current compression state.
Once the compressed DNN satisfies the size constraint, the evaluator will end the episode, and the accuracy evaluator will assess the pruned DNN's accuracy as an episode reward for the GNN-RL agent.
As opposed to the existing RL-based methods~\cite{he2018amc,yu2020agmc,liu2020AutoCompress}, with the DNN-Graph environment, the GNN-RL can learn to reach the desired model size in the reinforcement learning stage. Hence, it prevents us from adjusting pruning ratios and obtaining tiny compression ratios.
In the following, we will explain the details of the m-GNN and RL agent within our approach.


\subsection{Hierarchical Graph Representation}
\label{sec:hierarchicalgraph}
Computational graphs with their rich topological information may involve billions of operations~\cite{he2016ResNet}, making them bloated and hard to understand.
Nevertheless, such graphs often contain repetitive sub-graphs (a.k.a. motifs), such as 3$\times$3 convolutions or custom blocks in the state-of-the-art networks. 
We aim to simplify computational graphs by extracting the motifs and modeling them as hierarchical computational graphs.
Additionally, we coarsen the graph by replacing primitive operations such as \textit{add}, \textit{multiple}, and \textit{minus} with machine-learning high-level operations (e.g., convolution, pooling).

\begin{figure*}
\begin{center}
  \includegraphics[width=\textwidth]{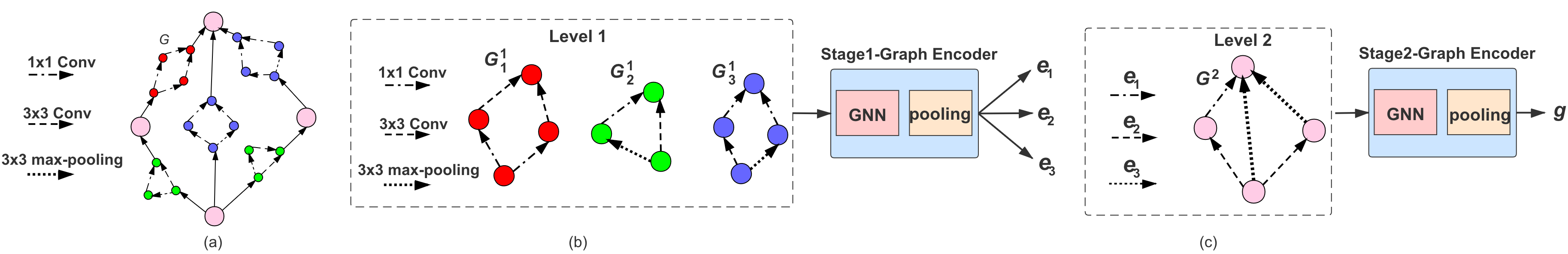}

\end{center}
   \caption{A two-level hierarchical computational graph and m-GNN. The sub-graphs are painted with red, blue, and green colors. }
\vskip -0.2in

\label{fig:1}
\end{figure*}

We formally model a given DNN as an $l$-level hierarchical computational graph, such that at the $l^{th}$ level (the top level), we would have the hierarchical computational graph set $\mathcal{G}^{l} = \{G^l\}$, where each item is a computational graph $G^l = (V^l,\mathcal{E}^l,\mathcal{G}^{l-1})$.
$V^l$ is the graph nodes corresponding to hidden states. $\mathcal{E}^l$ is the set of directed edges with a specific edge type associated with the operations. Lastly, $\mathcal{G}^{l-1} = \{G^{l-1}_0,G^{l-1}_1,...\}$ is the computational graph set at the $(l-1)$-level as well as the operation set at layer $l$. 
The hierarchical computation graph's size depends on the primitive operations we choose in $\mathcal{G}^{0}$. In this paper, we opted for commonly used machine-learning operations as the primitive operations for $\mathcal{G}^{0}$.
As an example, Figure \ref{fig:1} illustrates the idea behind generating hierarchical computational graphs using a sample graph $G$, where the edges are operations and the nodes are hidden states.
In the input graph, we choose three primitive operations $\mathcal{G}^{0} = $ \{1$\times$1 conv, 3$\times$3 conv, 3$\times$3 max-pooling\} corresponding to the three edge types. 
Then, we extract the repetitive subgraphs (i.e., $G^1_1$, $G^1_2$ and $G^1_3$), each denoting a compound operation, and decompose the graph $G$ into two hierarchical levels, as shown in Figure \ref{fig:1}~(b) and (c).

In practice, a 2-layer hierarchy is suitable for representing a DNN when our target is a convolutional layer. In the first layer, we represent the convolution operation and construct motifs, and in the second layer, we use the motifs to construct the DNN we aim to prune. 

\subsection{Multi-stage GNN}
Standard GNN and its variants~\cite{kipf2017gcn} are inherently flat~\cite{Ying2018DiffPool}. Since we model a given DNN as an $l-$level hierarchical computational graph, we propose a multi-stage GNN~(m-GNN), which embeds the hierarchical graph in $l$-stages according to its hierarchical levels and analyzes the motifs.
As depicted in Figure~\ref{fig:1}, m-GNN initially learns the lower level embeddings and uses them as the corresponding edge features in high-level computation graphs. Instead of learning node embeddings, m-GNN aims to learn the global graph representation. We further introduced a novel learning-based pooling strategy for every stage of embedding. With m-GNN, we only need embedding once for each motif on the computational graph. It is much more efficient and uses less memory than embedding a flat computation graph with standard GNN.

\textbf{Multi-stage embedding.} 
For the computational graphs $\mathcal{G}^{t} = \{G^{t}_0,G^{t}_1,...,G^{t}_{N_t}\}$ in the $t^{th}$ hierarchical level, we embed the computational graph $G^t_i = (V^t_i,\mathcal{E}^t_i,\mathcal{G}^{t-1}), i=\{1,2,...,N_t\}$ as:
\begin{equation}
    e^t_i = EncoderGNN_t(G^t_i, E_{t-1}) ,
\end{equation}
where $e^t_i$ is the embedding vector of $G^{t}_i$, $E_{t-1} = \{e^{t-1}_j\}, j=\{1,2,...,N_{t-1}\}$ is the embedding of the computational graphs at level ${t-1}$. We use $E_{t-1}$ as edge features at level $t$.
For level-1, $E_{0}$ contains the initial features (e.g., one-hot, and random standard) of the primitive operations $\mathcal{G}^{0}$ that we manually select. 
In the hierarchical computational graphs, each edge corresponds to a computational graph of the previous level and uses its graph embedding as the edge feature. Furthermore, the graphs at the same hierarchical level share the GNN's parameter.
At the top level ($l^{th}$ level) of the hierarchical graph $\mathcal{G}^{l} = \{G^l\}$, we only have one computational graph and its embedding is the DNN's final embedding $g$:
\begin{equation}
    g = EncoderGNN_l(G^l, E_{l-1})
\end{equation}

\begin{figure}%
\begin{center}
    \centerline
    {{\includegraphics[width=.4\linewidth]{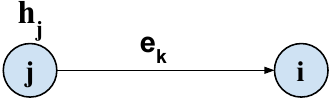} }}%
    \caption{Message passing from node j to node i.}%
    \label{fig:eq4}%
\end{center}
\vskip -0.3in
\end{figure}%
\textbf{Message passing.} In the multi-stage hierarchical embedding, we consider the edge features. However, in the standard graph convolutional networks (GCN)~\cite{kipf2017gcn}, it only passes the node features and the message passing function can be formulated as follows:
\begin{equation}
    h^{l+1}_i = \sum_{j\in N_i}\frac{1}{c_i}W^l h^l_j ,
\end{equation}
where $h$ is nodes' hidden states, $c_i$ is a constant coefficient, $N_i$ is node $i$ neighbors, and $W^l$ is GNN's learnable weight matrix.
Instead of standard message passing, in the multi-stage GNN, we add the edge features:
\begin{equation}
    h^{l+1}_i = \sum_{j\in N_i}\frac{1}{c_i}W^l (h^l_j\circ e^{l-1}_k) ,
\end{equation}
where $e^{l-1}_k$ is the features of edge $(i,j)$ and is also the embeddings of the $k^{th}$ graph at level $l-1$, such that the edge $(i,j)$ corresponds to the operation $G^{l-1}_k$. The operation $\circ$ denotes the element-wise product.

Many message-passing strategies exist, such as MP-GNN~\cite{gilmer2017neural}, in which they utilized a multi-layer perceptron.
In this work, assuming the scenario depicted in Figure~\ref{fig:eq4}, the message passing between $h_j$ and $e_k$ should essentially capture the amount of information in the node $j$~(i.e., the node feature $h_j$) that can flow to the node $i$ by using the factor $e_k$.
Thus, we selected element-wise product as the message passing function to assure that the same edge type can flow the same amount of information. Additionally, element-wise product satisfies the associative property in multi-stage message passing.

\textbf{Learning-based pooling.}
A typical GNN aims to learn the node embeddings of a graph~(e.g., learning node representation and perform node classification). However, our goal is to learn the graph representation of a given DNN. Thus, we introduced a learning-based pooling method for multi-stage GNN to learn the graph embedding from node embeddings. We define the graph embedding $e$ as:
\begin{equation}
    e = \sum_{i\in N}\alpha_i h_i + \sum_{j \in D}\alpha_j \vec{0},
\end{equation} 
where $N$ is the set of nodes, $h_i$ is the $i^{th}$ node embedding, $D$ is the set of pruned nodes, and $\alpha_i$ is the learnable weight coefficient.
In the multi-stage GNN, the computational graphs at the same hierarchical level share the GNN’s parameters, but in the pooling, each computational graph has its own learnable pooling parameters $\alpha$.
This parameter contains shared weights before and after pruning, and its dimension will not change. However, the number of nodes is not fixed after pruning, causing matrix dimension mismatch between the learning parameter $\alpha$ and the node features $h$~(i.e, $|\alpha| \neq |h|$).
As a remedy, we keep the size of $h$ constant by replacing the pruned node's features with a zero vector.

\textbf{Training.}
In GNN-RL, m-GNN is part of the actor-critic network inside the RL's policy network and will be updated end-to-end using the PPO algorithm~(see section \ref{sec:rloptimize} for detail).

\subsection{Network Pruning Using Reinforcement Learning and m-GNN}
We employed m-GNN together with reinforcement learning (RL) to find suitable network pruning strategies. In the following, we explain the details of the RL agent.

\label{sec:rloptimize}
\paragraph{Environment states.}
We use the generated hierarchical computational graph $\mathcal{G}^{l}$ for representing the DNN's state and the RL agent's environment state.
Since pruning the model causes its underlying graph topology to change, the DNN-Graph environment constantly updates the graph $\mathcal{G}^{l}$ after each pruning step to help the RL agent find the pruning policy on the current state.

\paragraph{Action space.}
The actions made by the RL agent are pruning ratios within a continuous space. Specifically, the GNN-RL agent's action space $A\in \mathbb{R}^{N \times 1}$, where $N$ is the number of pruning layers, is the pruning ratios for hidden layers: $A=[a_1, a_2, \dots, a_N]^T$, where $a_i \in [0,1)$ and $i = \{1,2,...,N\}$ is the pruning ratio for $i^{th}$ layer. 
GNN-RL agent makes the actions directly from the topology states:
\begin{equation}
    g = GraphEncoder(\mathcal{G}^{l}) ,
    \label{eq:graphencoder}
\end{equation}
\begin{equation}
    A = MLP(g) ,
    \label{eq:mlp}
\end{equation}
where $\mathcal{G}^{l}$ is the environment states, $g$ is the graph representation, and MLP is a multi-layer perceptron neural network. The graph encoder learns the topology embedding, and the MLP projects the embedding into hidden layers' pruning ratios. To bound the network's output within the action space, we apply a clamping (a.k.a. clipping) function within the range of [0,1) to the actions. 

\paragraph{Reward function.}
The reward function is $R_{err} = -Error$, where $Error$ is the compressed DNN's Top-1 error on the validation set. In computing the reward, we do not consider the model size, as the graph environment will automatically stop the search episode when the RL agent reaches the desired size.

\paragraph{RL policy.}
Various RL policies aim to search within a continuous action space, such as proximal policy optimization~(PPO)~\cite{schulman2017ppo} and deep deterministic policy gradient~(DDPG)~\cite{lillicrap2016ddpg}.
Although various state-of-the-art methods use the DDPG RL policy to search for the best pruning policy, we opted for the PPO RL policy, as it provided much better performance.
m-GNN is part of the actor-critic network inside the GNN-RL agent and will be updated end-to-end using the PPO algorithm.
Equation~\ref{eq:update} shows the objective function that we used in the PPO update policy. 
m-GNN is then trained by minimizing the objective function using the gradient descent algorithm.
\begin{equation}
\label{eq:update}
    L(\theta) = \hat{\mathbb{E}_t}[min(r_t(\theta)\hat{A}_t, clip(r_t(\theta),1-\epsilon,1+\epsilon)\hat{A}_t)],
\end{equation}
where $\theta$ is the policy parameter of the RL agent, which includes the m-GNN's parameters,
$\hat{\mathbb{E}_t}$ denotes the empirical expectation over time steps,
$r_t(\theta)$ is the ratio of the probability under the new and old policies,
$\hat{A}_t$ is the estimated advantage at time t, and $\epsilon$ is a clip hyperparameter, usually set to 0.1 or 0.2.

\section{Experimental Results}
To show the effectiveness of the GNN-RL, we evaluate our approach on various deep networks and compare our method with the following methods:
\begin{itemize}
    \item Traditional channel reduction methods, such as uniform empirical policies, SPP~\cite{wang2017SPP}, FP~\cite{Li2016handcraft}, RNP~\cite{Lin2017RNP}, FPGM~\cite{he2019FPGM}, SFP~\cite{he2018sfp}, DSA~\cite{ning2020dsa} and PFP~\cite{liebenwein2020pfp}.
    \item AutoML methods, such as NetAdapt~\cite{yang2018netadapt}, AutoPruner~\cite{luo2020AutoPruner}, EagleEye~\cite{li2020eagleeye}, AutoSlim~\cite{yu2019autoslim}, Meta-Pruning~\cite{liu2019metapruning}, AMC~\cite{he2018amc}, AGMC~\cite{yu2020agmc}, and random search (RS) with RL.
\end{itemize}

%

\subsection{Implementation Details}
\textbf{Graph representation settings.}
We model a given DNN as 2-layer hierarchical graphs, the primitive operations $\mathcal{G}^{0} = $ \{$k\times k$ conv, depth-wise conv, point-wise conv, $k\times k$ max-pooling\}.
When pruning, instead of using a one-hot vector, we initialized the hierarchical computational graph’s node and edge features using the standard distribution with a feature size of 20. 

\textbf{RL agent settings.}
We use the Adam optimizer to update the RL agent's actor-critic network, where the learning rate is $3~\times 10^{-4}$ and the $\beta = (0.9, 0.999)$. Moreover, we update the policy every $100$ search episodes, and in each updating round, we train the RL agent for 20 epochs. The discount factor is $\gamma = 0.99$, the clip parameter is $0.2$, and the standard deviation of actions is $0.5$.
Actor and critic networks contain a graph encoder with (hidden, embedding) size of (50,50) units and a multi-layer perceptron with a hidden size of 200 units. 
Although the output dimension of the actor-network is the number of actions, the critic-network’s MLP has only one output dimension.

\textbf{Dataset settings.}
The experiment involves multiple datasets, including CIFAR-10/100~\cite{Krizhevsky2009Cifar}, and ImageNet~\cite{Olga2015ImageNet}.
In the CIFAR-10/100 dataset, we sample $5K$ images from the test set as the validation set. 
In ImageNet~(ILSVRC-2012), we split $10K$ images from the test set as the validation set. When searching, the DNN-Graph environment uses the compressed model's $R_{err}$ on the validation set as the GNN-RL agent's reward. 

\textbf{Network settings.}
Since ResNet contains residual connections between convolutional layers, different pruning ratio among residual connected layers leads to feature maps mismatch. Instead of removing the residual connections, we share the pruning ratio between residual connected layers~(i.e., equal pruning ratio in residual connected layers).
For MobileNet networks, applying regular filter pruning on depth-wise/point-wise convolution layers causes information loss. Thus, instead of pruning such filters separately, we only prune linear expansion layers and point-wise filters within MobileNet blocks.
Since residual connections are between linear expansion layers in MobileNet-v2, we share the linear expansion layers' pruning ratio.
Lastly, to prune ShuffleNet networks, we consider its blocks together and perform channel pruning inside the blocks. In a ShuffleNet block, we do not prune the expansion layer (the output layer of the block), which can preserve the number of output channels.

\begin{figure}[ht]
\begin{center}
\centerline{\includegraphics[width=\columnwidth]{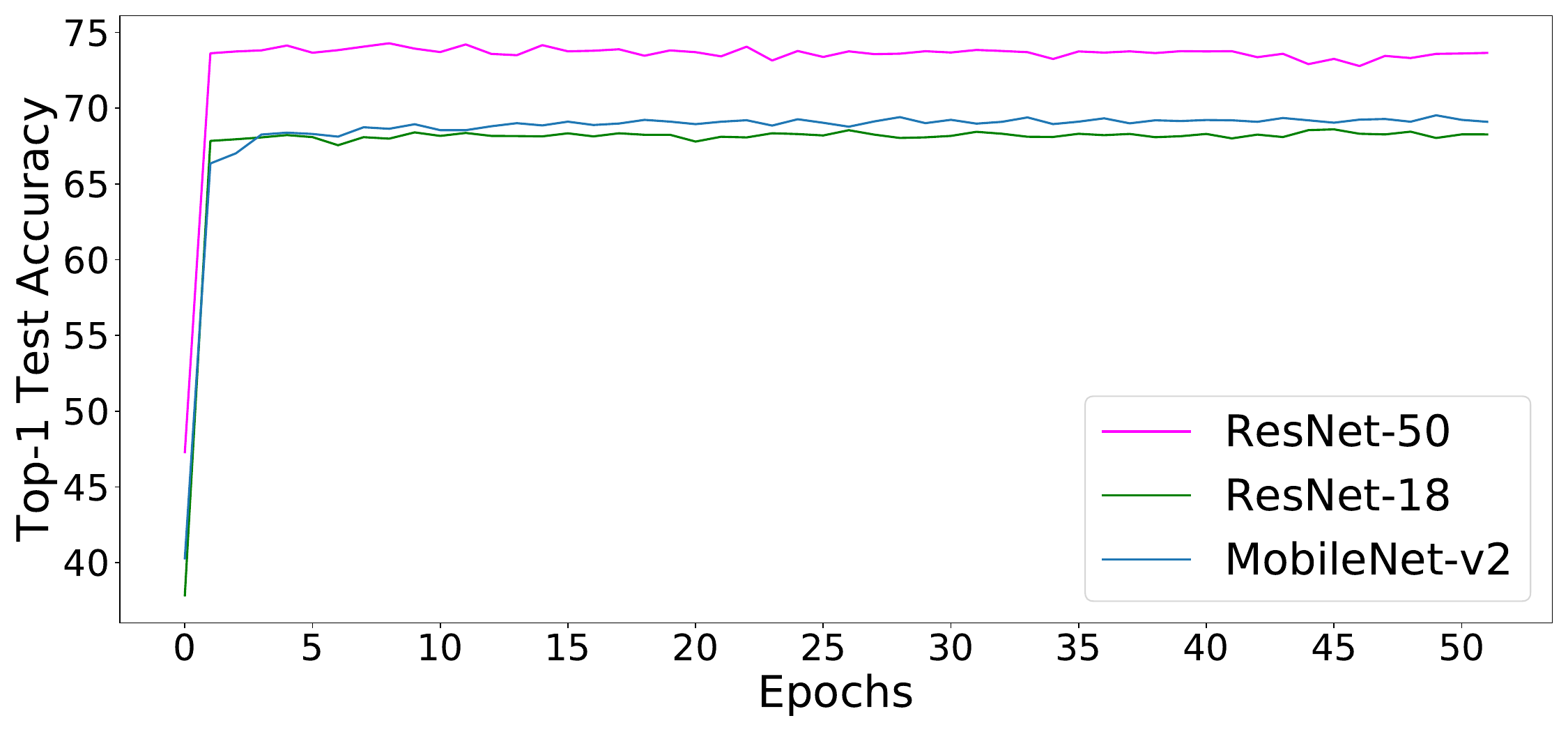}}
 \vskip -0.1in
\caption{Accuracy recovery after different fine-tuning epochs.}
\label{fig:acc_epochs}
\end{center}
 \vskip -0.3in
\end{figure}

\textbf{Fine-tuning settings.}
As we observed a direct correlation between the pre-/post-fine-tuning accuracy, when pruning on CIFAR-10/100, we perform fine-tuning after pruning. However, the validation accuracy on the ImageNet dataset is sensitive to the compression ratio, particularly for the MobileNet-v1/2. Without fine-tuning, high compression ratios lead to a considerable accuracy drop. Thus, on MobileNet-v1/2 trained on ImageNet, we perform one additional fine-tuning epoch before we compute the reward to ensure that the RL agent gets a valuable reward~(as depicted in Figure~\ref{fig:acc_epochs}, one epoch of fine-tuning can recover the majority of accuracy loss).
After pruning, a 150-epochs fine-tuning process is applied to the pruned DNNs on ImageNet/CIFAR-100, and a 100-epochs fine-tuning is applied on CIFAR-10.
We use the SGD optimizer, where the $\alpha=5~\times 10^{-3}$, $B=512$, and the weight decay is $5~\times 10^{-4}$. In each epoch, the cosine learning rate decay is applied.

\begin{table}[t]
\caption{Top-1 accuracy results for pruned ResNets on CIFAR-10 and ShuffleNet on CIFAR-100. $\Delta $Top-1 shows the top-1 accuracy gap between the pruned and the original model.}
\label{tab:res_cifar}
\begin{center}
\begin{small}

\begin{tabular}{lccccccr}
\toprule
Model   & Method & FLOPs $\downarrow$  &Top-1 & $\Delta $Top-1 \\
\midrule
\multirow{3}{*}{Res110}    &AGMC   & $50\%$    & $ 93.08$     & $-0.60 $\\
                              &RS     & $50\%$ &  $87.26 $    & $ -6.42$ \\
                              &PFEC   & $39\%$    & $ 93.30$     & $-0.20 $\\
                              &SFP   & $41\%$    & $ 93.38$     & $+0.16 $\\
                              &FPGM   & $52\%$    & $ 93.74$     & $-0.60 $\\
                              &{GNN-RL} & $52\%$  & $ {94.31}$ & $ {+0.63}$ \\
\midrule
\multirow{5}{*}{Res56}     &Uniform& $50\%$   & $87.5$ & $-5.89$ \\
                           &AMC    & $50\%$  & $90.20$ & $-3.19$ \\
                           &FPGM   & $59\%$  & $93.26$ & $-0.02$ \\                           
                           &AGMC   & $50\%$  & $92.00$ & $-1.39$\\
                           &EagleEye   & $50\%$  & $94.66$ & N/A\\
                           &{GNN-RL} & $54\%$ & ${93.49}$ & $ {+0.10}$ \\
\midrule
\multirow{3}{*}{Res32}     &AGMC   & $50\%$ &  $90.96$ & $-1.67$ \\
                           &RS     & $50\%$ &  $ 89.57$ & $-3.06$ \\
                           &SFP     & $42\%$ &  $ 92.08$ & $-0.55$ \\
                           &FPGM     & $50\%$ &  $ 91.93$ & $-0.70$ \\
                           &{GNN-RL} & ${51}$\% & ${92.58}$ & ${-0.05}$ \\
\midrule
\multirow{5}{*}{Res20}    
                          &Uniform& $50\%$ &  $84.00$ & $-7.73$ \\
                          &AMC    & $50\%$ &  $86.40$ & $-5.33$ \\
                          &AGMC   & $50\%$ &  $88.42$ & $-3.31$ \\
                          &SFP   & $42\%$ &  $90.83$ & $-1.37$ \\
                          &FPGM   & $42\%$ &  $91.09$ & $-1.11$ \\
                          &DSA   & $50\%$ &  $91.38$ & $-0.79$ \\
                          & {GNN-RL} & $ {51\%}$   &  ${91.31}$ &  {-0.42} \\

\midrule
\multirow{3}{*}{Shuff-v1}  &AGMC   &  $40\%$ & $65.26 $     & $ -3.38$\\
                           &RS     &  $40\%$ &  $63.70 $    & $ -4.94$ \\
                           & {GNN-RL} & $ {42\%} $ & $  {67.10} $ & $  {-2.84} $ \\
\midrule
\multirow{3}{*}{Shuff-v2} 
                           &AGMC    & $40\%$  & $ 66.28$ & $-2.57 $ \\
                           &RS   &  $40\%$  & $ 65.74$ & $ -3.11$\\
                           & {GNN-RL} & $ {46\%} $  & $ {66.64}$ & $ {-2.21}$ \\
                            
\bottomrule
\end{tabular}

\end{small}

\end{center}
\vspace{-15pt}
\end{table}

\begin{table}[t]
\caption{Top-1 accuracy results for pruned models on ImageNet. $\Delta $Top-1 shows the top-1 accuracy gap }
\label{tab:res_imagenet}

\begin{center}
\begin{small}
\begin{tabular}{lccccccr}
\toprule
Model  & Method & FLOPs $\downarrow$  & Top-1 & $\Delta$ Top-1 \\
\midrule
\multirow{4}{*}{VGG16}        &FP & $80\%$  & $55.90$ & $-14.6$ \\
                               &RNP & $80\%$ & $66.92$ & $-3.58$ \\
                               &SPP & $80\%$ & $68.20$ & $-2.30$ \\
                             &AMC & $80\%$   & $69.10$ & $-1.40$ \\
                             &AutoPruner & $74\%$   & $69.20$ & $-2.39$ \\
                               &  {GNN-RL} & 80\% & $   {70.99}$ & $  {+0.49}$ \\
\midrule
\multirow{4}{*}{Res18}        & FPGM &  $42\%$  & 68.41  & -1.87  \\
                               &  SFP&    $42\%$  & 67.10  & -3.18 \\
                                &  PFP-B    & $43\%$  & 65.65 & -4.09 \\
                                &  {GNN-RL} &     {51\%}   &   {68.66}  &  {-1.10}   \\
\midrule
\multirow{5}{*}{Res50}        & AutoPruner &   $44 \%$ & 73.84  & -2.31  \\
                               & Meta-Pruning&   $50 \%$  & 73.40  &  -3.20  \\
                               & AutoSlim &   $ 50 \%$ &   74.00 & -2.10\\
                              
                             &EagleEye        &  $50 \%$  & 74.20& N/A \\
                               & {GNN-RL} &   {53\%} &  {74.28}  &  {-1.82} \\

\midrule
\multirow{7}{*}{MB-v1}     &Uniform   &  $25\%$  &  $68.40 $ & $ -2.20 $ \\
                              &NetAdapt   &  $34\%$     &  $69.10 $ & $  -1.50$ \\
                              &AMC     &  $34\%$        &  $ 70.50$ & $ -0.40$ \\
                              &Meta-Pruning     &  $35\%$ &  $ 70.60$ & $ 0$ \\
                              &EagleEye &  $25\%$  &  $70.90 $ & N/A \\
                              &{GNN-RL} & $30\%$  &$70.70$ & $-0.20 $ \\
                              &AMC     &  $60\%$  &  $ 68.90$ & $ -2.00$ \\
                              &  {GNN-RL} & ${ 60\%}$ &  $  { 69.50}$ & $   {-1.40}$ \\
\midrule
\multirow{3}{*}{MB-v2}     &AMC   & $27\%$   &  $70.80 $ & $ -1.00$ \\
                              &Meta-Pruning     &  $27\%$  &  $71.20 $ & $ -3.50$ \\
                              &NetAdapt     &  $25\%$  &  $70.00 $ &  N/A \\
                            
                              &{GNN-RL} &  $   {42\%}  $  & $70.04 $ & $ -1.83$ \\

\bottomrule
\end{tabular}
\end{small}
\end{center}
\vspace{-15pt}

\end{table}


\begin{figure}[ht]
\begin{center}
\centerline{\includegraphics[width=\columnwidth]{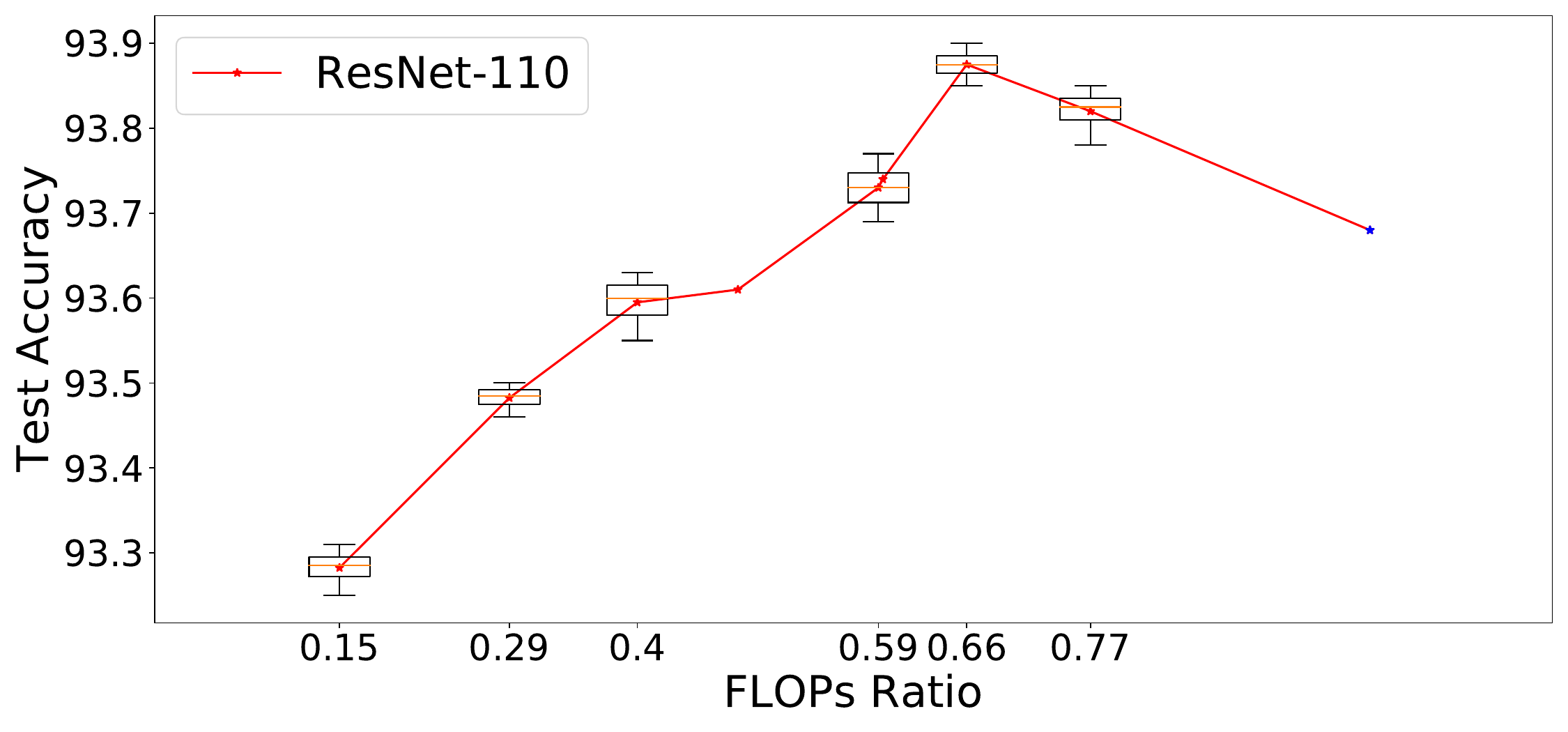}}
\caption{Test accuracy of ResNet-110 using various FLOPs ratios.}
\label{fig:5}
\end{center}
\end{figure}

\subsection{Comparisons with State-of-the-art}

Tables~\ref{tab:res_cifar} and~\ref{tab:res_imagenet} compare the pruning efficiency of GNN-RL with state-of-the-art methods trained on CIFAR-10/100 and ImageNet datasets.
Results show that GNN-RL achieves competitive results compared with state-of-the-art methods and even  outperforms many of them either by higher accuracy or pruning ratio. 
Pruning ResNet-110/56 models even caused higher test accuracy than the original model, which could be due to over-fitting, as the accuracy on the training set was 100\%. To verify our assumption, we explored the relationship between the FLOPs constraints and the accuracy. 
Figure~\ref{fig:5} shows that the 66\%-FLOPs Resnet-110 can get the highest test accuracy. When the FLOPs reduction ratio exceeds 0.66, the test accuracy drops intensively.



\begin{figure*}
\begin{center}
  \includegraphics[width=\linewidth]{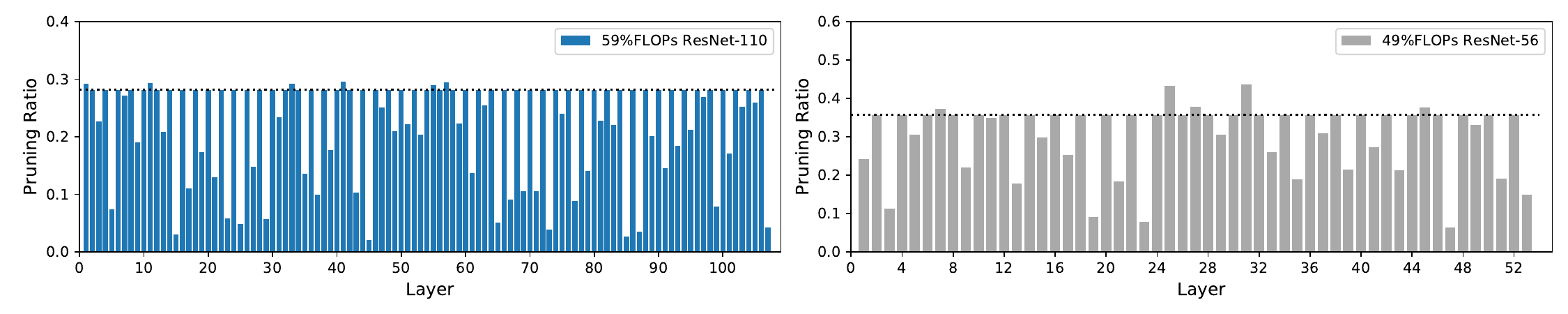}

\end{center}
   \caption{The hidden layers' pruning ratio of $59\%$ FLOPs ResNet-110 and $49\%$ FLOPs ResNet-56. The bars that tangent with the dot-line are the residual connection layers. 
   }

\label{fig:4}
\end{figure*}
We further analyzed the redundancy and the importance of ResNet layers. Figure~\ref{fig:4} shows the hidden layers' pruning ratios on ResNet-110 and ResNet-56. 
GNN-RL agent automatically learns that the residual connection layers with ResNet are redundant. Thus, it applies more pruning on such layers.
Moreover, GNN-RL's results are inconsistent with previous handcrafted pruning works, which assume that the deeper layers are more important for final predictions and tend to prune less. However, GNN-RL applies more pruning to the middle (layers 45 to 65) and deep (layers 90 to 109) layers within ResNet-110. Such an observation proves that handcrafted rules are not generalizable to all DNNs, particularly those with residual connections.

\subsection{Ablation Study}
\paragraph{Recoverability.}
A noteworthy feature of the models pruned by GNN-RL is that they can rapidly recover from accuracy loss. Figure~\ref{fig:acc_epochs} demonstrates the fine-tuning learning curve for ResNet-50/18 and MobileNet-v2.
We noticed that after only one epoch of fine-tuning, we can recover from the accuracy loss caused by pruning.
We only needed less than 50 epochs for further accuracy improvement, which is far less than the number of fine-tuning epochs used by other state-of-the-art methods.

\paragraph{Effectiveness of reinforcement learning policy.}
\label{sec:rl}

\begin{figure}[t]
\begin{center}
\centerline{\includegraphics[width=\columnwidth]{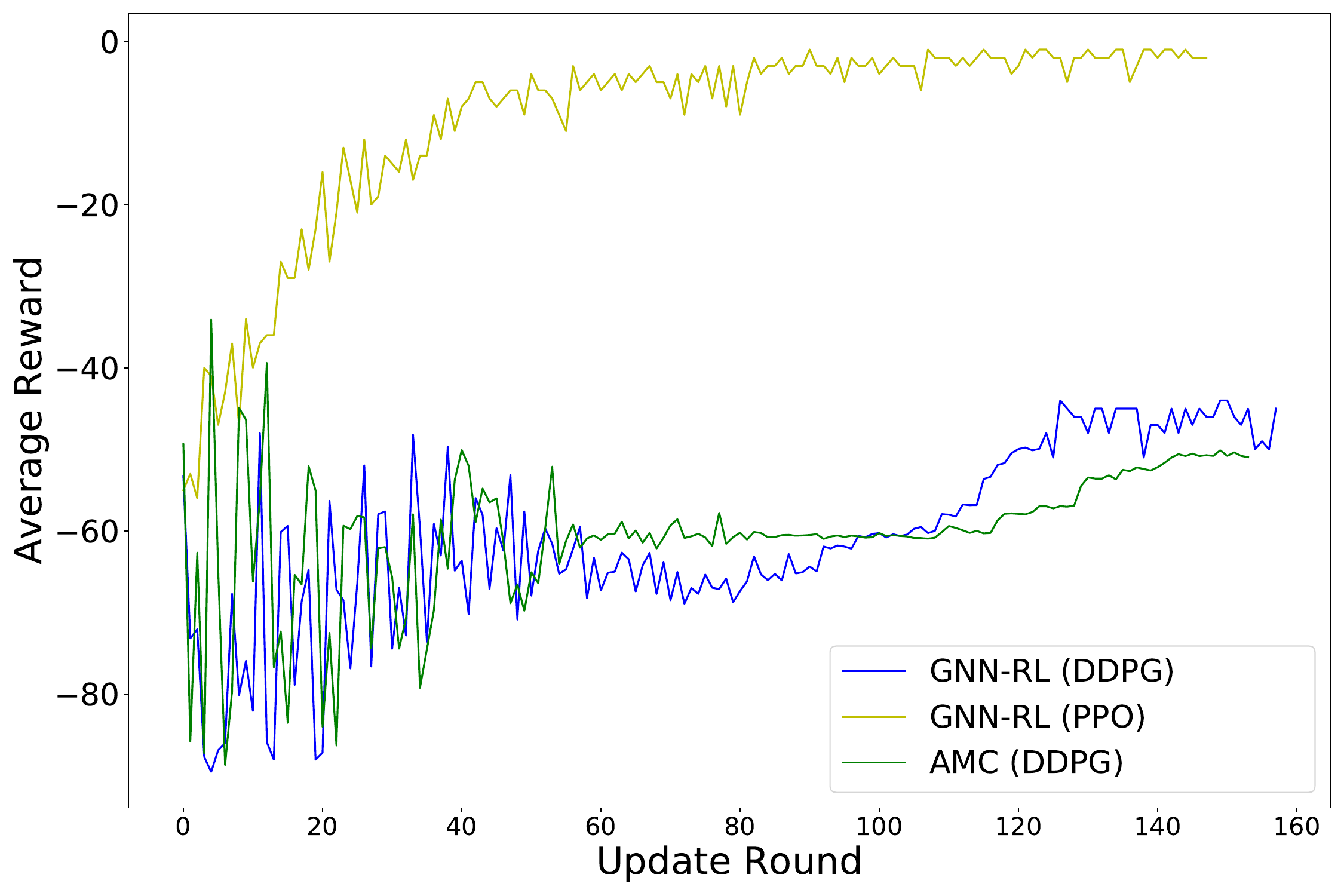}}
\caption{Learning curve of RL policies; DDPG vs. PPO.}
\label{fig:reward_rounds}
\end{center}
\end{figure}

To the best of our knowledge, existing RL-based pruning methods (e.g., AMC) employ the DDPG policy for implementing their RL agent.
However, we found out that PPO policy offers better performance and converges faster than DDPG.
Figure~\ref{fig:reward_rounds} shows the RL agents' learning curve, comparing GNN-RL implemented with DDPG and PPO with AMC.
In addition to better performance, PPO has less tunable hyperparameters, making PPO easier to configure and tune. 

Additionally, the RL policy network of GNN-RL is a tiny neural network that contains a graph encoder and a two-layer perceptron with limited action and environment space, leading to efficient training and inference. For instance, in the experiment conducted on CIFAR-10, GNN-RL could converge within half a GPU hour using an Nvidia V100.


\paragraph{Topology transferability.}
\begin{figure}[t]
\begin{center}
\centerline{\includegraphics[width=\columnwidth]{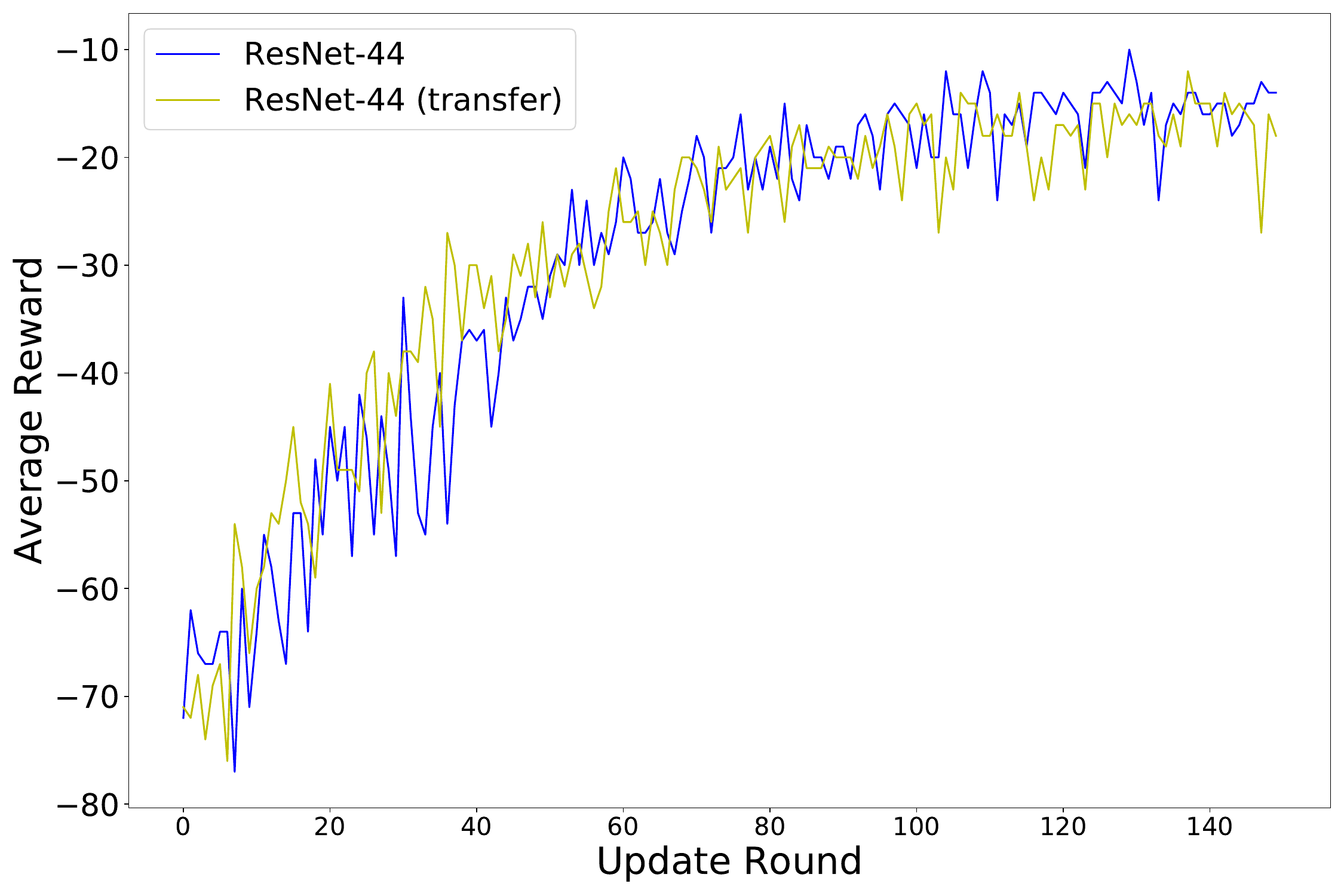}}
\caption{Topology transferability.}
\label{fig:topology_transfer}
\end{center}
\end{figure}

The topology transferability is a key factor to demonstrate whether GNN embeddings are necessary or even applicable. We aim to prove that GNN-RL can learn a transferable policy from a given DNN topology. 
Intuitively, GNNs trained on a topology can be transferred to a simpler topology. Thus, we first trained GNN-RL on ResNet-56 and then transferred the graph encoder to ResNet-44.
When searching for the pruning policy, we disabled the graph encoder's gradients and only updated the MLP component, which projects the topology embeddings into the action space. 
Figure~\ref{fig:topology_transfer} shows the RL's learning curve for direct pruning search and the transferred policy.
We noticed that the transferred GNN-RL has a similar learning curve to direct search on ResNet-44, indicating that the learned policy can be reused for networks with similar topology. Such a feature offers a rapid pruning process~($1.12\times$ faster for each round) with much less computing time, as we only need to update the MLP's parameter. Under the transfer policy, we achieved $93.23\%$ accuracy with 51\% FLOPs reduction comparable to the original ResNet-44, which is 93.10\%.\\

\begin{table}
\begin{center}
\small

\caption{ResNet-110 node classification results using GCN and m-GNN.\label{tab:node_classif}
}

\begin{tabular}{|l|c|c|c|c|}
\hline
Graph &  Nodes &Edges & Method &Acc.\% \\
\hline\hline
{Plain}  &394,412 & 581,332   &  GCN& 83.50   \\
{Hierarchical}  &12,460 &   20,221 &  m-GNN& 84.20    \\

\hline
\end{tabular}
\end{center}
\vspace{-10pt}
\end{table}
 \vskip -0.2in


\begin{table}[t]

\begin{center}
\begin{small}
\begin{sc}
\caption{The latency and GPU memory usage before and after pruning.
}

\label{table_3}
\begin{tabular}{lcccr}

Model & FLOPs & Latency &GPU Mem. \\

\midrule

\multirow{2}{*}{VGG-16}          & $100\%$            &  $0.11ms$&       $528$MB \\
                                    & $20\%$          &  $0.08ms$&       $387$MB \\

\midrule

\multirow{2}{*}{ResNet-110}          & $100\%$            & $1.04ms$&  $6.9$MB      \\
                                    & $48\%$         &$0.98ms$& $3.4$MB\\
                                    
\midrule

\multirow{2}{*}{ResNet-56}          & $100\%$            & $0.52ms$   &       $3.4$MB \\
                                    & $46\%$             &$0.43ms$    &    $1.7$MB \\
\midrule

\multirow{2}{*}{ResNet-44}          & $100\%$            & $0.37ms$   &       $2.7$MB \\
                                    & $49\%$             &$0.34ms$    &    $1.3$MB \\

\midrule
\multirow{2}{*}{ResNet-32}   & $100\%$             & $0.33ms$&       $1.9$MB \\
                                    & $49\%$          &$0.26ms$& $942$KB \\
                                    
\midrule
\multirow{2}{*}{ResNet-20}   & $100\%$             & $0.20ms$&       $1.1$MB \\
                                    & $49\%$          &$0.16ms$& $548$KB \\
\midrule
\multirow{2}{*}{MobileNet-v1}          & $100\%$           & $0.22ms$&       $13$MB \\
                                    & $79\%$          &$0.19ms$& $7.5$MB \\
\midrule
\multirow{2}{*}{MobileNet-v2}          & $100\%$           & $0.34ms$&       $9.3$MB \\
                                    & $79\%$          &$0.32ms$& $ 7.4$MB \\
\midrule
\multirow{2}{*}{ShuffleNet-v1}        & $100\%$           & $0.45ms$&       $4.1$MB \\
                                    & $58\%$          &$0.43ms$& $ 2$MB \\
                                    \midrule
\multirow{2}{*}{ShuffleNet-v2}          & $100\%$           & $0.51ms$&       $5.4$MB \\
                                    & $54\%$          &$0.50$ms& $ 2.7$MB \\

\bottomrule

\end{tabular}%
\end{sc}
\end{small}
\end{center}
\end{table}

\paragraph{The impact of hierarchical graph representation (m-GNN).}
We realized that motifs frequently appearing in computation graphs should have the same embedding. With m-GNN, we only need to embed each motif once. However, a plain GNN disregards this hypothesis, leading to repetitive embeddings and incurring unnecessary costs.
To further prove our hypothesis, we experimented with the ResNet-110's computational graph and labeled the nodes according to their hidden layer. Table~\ref{tab:node_classif} shows that hierarchical representation causes to have smaller graph sizes and helps our method to achieve a higher node classification accuracy with fewer graph nodes and edges, saving a considerable amount of computing resources.
Moreover, using m-GNN led to a faster convergence and as shown in Tables~\ref{tab:res_cifar} and~\ref{tab:res_imagenet} resulted into more precise pruned models.

\subsection{Inference Acceleration and Memory Saving}
The inference and memory usage of compressed DNNs are essential metrics to determine the possibility of DNN deployment on a given platform.
Thus, we evaluated the pruned models' inference latency using PyTorch 1.7.1 on an Nvidia GTX 1080Ti GPU and recorded the GPU memory usages. The ResNet-110/56/44/32/20, VGG-16, and MobileNet/ShuffleNet networks are evaluated on the CIFAR-10, ImageNet, and CIFAR-100 datasets with batch size 32.

Table~\ref{table_3} shows the inference accelerations and memory savings on our GPU. All the models pruned by GNN-RL achieve noteworthy inference acceleration and GPU memory reductions. Particularly, for the VGG-16, the original model's GPU memory usage is 528 MB since it has a very compact dense layer, which contributes little to FLOPs but leads to extensive memory requirement. 
The GNN-RL prunes convolutional layers and significantly reduces the feature map size, consuming 141 MB less memory than the original version. The inference acceleration on VGG-16 is also noticeable, with $1.38 \times$ speed up on the ImageNet.

The inference acceleration for mobile-friendly DNNs may seem relatively insignificant. However, such models are designed for deployment on mobile devices. Thus, we believe that our tested GPU, with its extensive resources, does not take advantage of the mobile-friendly properties.

\section{Conclusion}
This paper proposed a neural network pruning method called GNN-RL that utilizes graph neural networks and reinforcement learning to exploit a topology-aware compression policy. We introduced the DNN-Graph environment that converts compression states to a topology modification process and allows GNN-RL to learn the desired compression ratio without human intervention. To efficiently embed DNNs and take advantage of motifs, we introduced m-GNN, a new multi-stage graph embedding method.
In our experiments, GNN-RL is validated and verified on over-parameterized and mobile-friendly networks. For over-parameterized models pruned by GNN-RL, ResNet-110/56, the test accuracy even outperformed the original models, i.e. $+0.63\%$ on ResNet-110 and $+0.1\%$ on ResNet-56. 
For mobile-friendly DNNs, the $40\%$ FLOPs MobileNet-v2 pruned by GNN-RL with only $1.4\%$ test accuracy loss. Additionally, all the pruned models accelerated the inference speed and saved a considerable amount of memory usage. Most importantly, GNN-RL learns the topology transfer policy, enabling the GNN-RL to prune various DNNs with transfer learning.
\section*{Acknowledgement}
We would like to thank the research IT team of Iowa State University for their continuous support in conducting the experiments. Experiments presented in this paper were carried out on the Pronto GPU cluster at ISU. 
This research has been supported by publication award of the computer science department at Iowa State University. Furthermore, we received support from Software Campus through the German Federal Ministry of Education and Research (BMBF), and the state of Hesse as part of the NHR Program.


\bibliography{example_paper}

\begin{thebibliography}{62}
\providecommand{\natexlab}[1]{#1}
\providecommand{\url}[1]{\texttt{#1}}
\expandafter\ifx\csname urlstyle\endcsname\relax
  \providecommand{\doi}[1]{doi: #1}\else
  \providecommand{\doi}{doi: \begingroup \urlstyle{rm}\Url}\fi

\bibitem[Anwar et~al.(2017)Anwar, Hwang, and Sung]{Anwar2017Structured}
Anwar, S., Hwang, K., and Sung, W.
\newblock Structured pruning of deep convolutional neural networks.
\newblock \emph{Proc. of the J. Emerg. Technol. Comput. Syst.}, 13\penalty0
  (3), February 2017.
\newblock ISSN 1550-4832.

\bibitem[Boyd et~al.(2011)Boyd, Parikh, Chu, Peleato, and
  Eckstein]{Boyd2011ADMM}
Boyd, S., Parikh, N., Chu, E., Peleato, B., and Eckstein, J.
\newblock Distributed optimization and statistical learning via the alternating
  direction method of multipliers.
\newblock \emph{Found. Trends Mach. Learn.}, 3\penalty0 (1):\penalty0 1–122,
  January 2011.
\newblock ISSN 1935-8237.

\bibitem[Chatzianastasis et~al.(2021)Chatzianastasis, Dasoulas, Siolas, and
  Vazirgiannis]{chatzianastasis2021graph}
Chatzianastasis, M., Dasoulas, G., Siolas, G., and Vazirgiannis, M.
\newblock Graph-based neural architecture search with operation embeddings.
\newblock In \emph{Proc. of the IEEE/CVF International Conference on Computer
  Vision (CVPR)}, pp.\  393--402, 2021.

\bibitem[Chen et~al.(2020)Chen, Chen, and Pan]{chen2020storage}
Chen, J., Chen, S., and Pan, S.~J.
\newblock Storage efficient and dynamic flexible runtime channel pruning via
  deep reinforcement learning.
\newblock In Larochelle, H., Ranzato, M., Hadsell, R., Balcan, M.~F., and Lin,
  H. (eds.), \emph{Advances in Neural Information Processing Systems},
  volume~33, pp.\  14747--14758. Curran Associates, Inc., 2020.

\bibitem[Chen et~al.(2021)Chen, Ji, Ding, Fang, Wang, Zhu, Liang, Shi, Yi, and
  Tu]{chen2021oto}
Chen, T., Ji, B., Ding, T., Fang, B., Wang, G., Zhu, Z., Liang, L., Shi, Y.,
  Yi, S., and Tu, X.
\newblock Only train once: A one-shot neural network training and pruning
  framework.
\newblock In Ranzato, M., Beygelzimer, A., Dauphin, Y., Liang, P., and Vaughan,
  J.~W. (eds.), \emph{Advances in Neural Information Processing Systems},
  volume~34, pp.\  19637--19651. Curran Associates, Inc., 2021.

\bibitem[Chin et~al.(2020)Chin, Ding, Zhang, and Marculescu]{chin2020legr}
Chin, T.-W., Ding, R., Zhang, C., and Marculescu, D.
\newblock Towards efficient model compression via learned global ranking.
\newblock In \emph{Proc. of the IEEE/CVF Conference on Computer Vision and
  Pattern Recognition (CVPR)}, June 2020.

\bibitem[Dudziak et~al.(2021)Dudziak, Chau, Abdelfattah, Lee, Kim, and
  Lane]{Dudziak2021BPR_NAS}
Dudziak, L., Chau, T., Abdelfattah, M.~S., Lee, R., Kim, H., and Lane, N.~D.
\newblock {BRP-NAS}: Prediction-based {NAS} using gcns, 2021.

\bibitem[Gao et~al.(2021)Gao, Huang, Cai, and Huang]{gao_network_2021}
Gao, S., Huang, F., Cai, W., and Huang, H.
\newblock Network {Pruning} via {Performance} {Maximization}.
\newblock In \emph{Proc. of the {IEEE}/{CVF} {Conference} on {Computer}
  {Vision} and {Pattern} {Recognition} ({CVPR})}, pp.\  9270--9280, June 2021.

\bibitem[Gholami et~al.(2021)Gholami, Kim, Dong, Yao, Mahoney, and
  Keutzer]{gholami2021survey_quantization}
Gholami, A., Kim, S., Dong, Z., Yao, Z., Mahoney, M.~W., and Keutzer, K.
\newblock A survey of quantization methods for efficient neural network
  inference, 2021.

\bibitem[Gilmer et~al.(2017)Gilmer, Schoenholz, Riley, Vinyals, and
  Dahl]{gilmer2017neural}
Gilmer, J., Schoenholz, S.~S., Riley, P.~F., Vinyals, O., and Dahl, G.~E.
\newblock Neural message passing for quantum chemistry.
\newblock In \emph{Proc. of International conference on machine learning}, pp.\
   1263--1272. PMLR, 2017.

\bibitem[Guo et~al.(2020)Guo, Wang, Li, and Yan]{guo2020dmcp}
Guo, S., Wang, Y., Li, Q., and Yan, J.
\newblock Dmcp: Differentiable markov channel pruning for neural networks.
\newblock In \emph{Proc. of the IEEE/CVF Conference on Computer Vision and
  Pattern Recognition (CVPR)}, June 2020.

\bibitem[Guo et~al.(2016)Guo, Yao, and Chen]{Guo2016unstructured}
Guo, Y., Yao, A., and Chen, Y.
\newblock Dynamic network surgery for efficient dnns.
\newblock In Lee, D., Sugiyama, M., Luxburg, U., Guyon, I., and Garnett, R.
  (eds.), \emph{Proc. of the Advances in Neural Information Processing
  Systems}, volume~29, pp.\  1379--1387. Curran Associates, Inc., 2016.

\bibitem[Guo et~al.(2019)Guo, Zheng, Tan, Chen, Chen, Zhao, and
  Huang]{Guo2019NAS_NAT}
Guo, Y., Zheng, Y., Tan, M., Chen, Q., Chen, J., Zhao, P., and Huang, J.
\newblock {NAT}: Neural architecture transformer for accurate and compact
  architectures.
\newblock In \emph{Proc. of the Advances in Neural Information Processing
  Systems}, volume~32, pp.\  737--748. Curran Associates, Inc., 2019.

\bibitem[Han et~al.(2016)Han, Mao, and Dally]{han2015deep}
Han, S., Mao, H., and Dally, W.~J.
\newblock Deep compression: Compressing deep neural networks with pruning,
  trained quantization and huffman coding.
\newblock In \emph{Proc. of International Conference on Learning
  Representations (ICLR)}, 2016.

\bibitem[He et~al.(2016)He, Zhang, Ren, and Sun]{he2016ResNet}
He, K., Zhang, X., Ren, S., and Sun, J.
\newblock Deep residual learning for image recognition.
\newblock In \emph{Proc. of the IEEE conference on computer vision and pattern
  recognition}, pp.\  770--778, 2016.

\bibitem[He et~al.(2017)He, Zhang, and Sun]{he2017handcraft_channel}
He, Y., Zhang, X., and Sun, J.
\newblock Channel pruning for accelerating very deep neural networks.
\newblock In \emph{Proc. of the IEEE International Conference on Computer
  Vision}, pp.\  1389--1397, 2017.

\bibitem[He et~al.(2018{\natexlab{a}})He, Kang, Dong, Fu, and Yang]{he2018sfp}
He, Y., Kang, G., Dong, X., Fu, Y., and Yang, Y.
\newblock Soft filter pruning for accelerating deep convolutional neural
  networks.
\newblock In \emph{International Joint Conference on Artificial Intelligence
  (IJCAI)}, pp.\  2234--2240, 2018{\natexlab{a}}.

\bibitem[He et~al.(2018{\natexlab{b}})He, Lin, Liu, Wang, Li, and
  Han]{he2018amc}
He, Y., Lin, J., Liu, Z., Wang, H., Li, L.-J., and Han, S.
\newblock {AMC}: {AutoML} for model compression and acceleration on mobile
  devices.
\newblock In \emph{Proc. of the European Conference on Computer Vision (ECCV)},
  pp.\  784--800, 2018{\natexlab{b}}.

\bibitem[He et~al.(2019)He, Liu, Wang, Hu, and Yang]{he2019FPGM}
He, Y., Liu, P., Wang, Z., Hu, Z., and Yang, Y.
\newblock Filter pruning via geometric median for deep convolutional neural
  networks acceleration.
\newblock In \emph{Proc. of the IEEE Conference on Computer Vision and Pattern
  Recognition (CVPR)}, 2019.

\bibitem[Hinton et~al.(2015)Hinton, Vinyals, and Dean]{hinton2015distilling}
Hinton, G., Vinyals, O., and Dean, J.
\newblock Distilling the knowledge in a neural network.
\newblock In \emph{Proc. of NIPS Deep Learning and Representation Learning
  Workshop}, 2015.

\bibitem[Howard et~al.(2019)Howard, Sandler, Chu, Chen, Chen, Tan, Wang, Zhu,
  Pang, Vasudevan, et~al.]{howard2019searching}
Howard, A., Sandler, M., Chu, G., Chen, L.-C., Chen, B., Tan, M., Wang, W.,
  Zhu, Y., Pang, R., Vasudevan, V., et~al.
\newblock Searching for mobilenetv3.
\newblock In \emph{Proc. of the IEEE/CVF International Conference on Computer
  Vision}, pp.\  1314--1324, 2019.

\bibitem[Huang et~al.(2018)Huang, Liu, Van~der Maaten, and
  Weinberger]{huang2018condensenet}
Huang, G., Liu, S., Van~der Maaten, L., and Weinberger, K.~Q.
\newblock Condensenet: An efficient densenet using learned group convolutions.
\newblock In \emph{Proc. of the IEEE conference on computer vision and pattern
  recognition}, pp.\  2752--2761, 2018.

\bibitem[Jian-Hao~Luo(2020)]{luo2020AutoPruner}
Jian-Hao~Luo, J.~W.
\newblock Autopruner: An end-to-end trainable filter pruning method for
  efficient deep model inference.
\newblock \emph{Pattern Recognition}, 2020.

\bibitem[Kipf \& Welling(2017)Kipf and Welling]{kipf2017gcn}
Kipf, T.~N. and Welling, M.
\newblock Semi-supervised classification with graph convolutional networks.
\newblock In \emph{Proc. of the International Conference on Learning
  Representations (ICLR)}, 2017.

\bibitem[Krizhevsky \& Hinton(2009)Krizhevsky and Hinton]{Krizhevsky2009Cifar}
Krizhevsky, A. and Hinton, G.
\newblock Learning multiple layers of features from tiny images.
\newblock 2009.

\bibitem[Lai et~al.(2021)Lai, Zhang, Liu, Chang, Liao, Chuang, Qian, Khurana,
  Cox, and Glass]{lai2021parp}
Lai, C.-I.~J., Zhang, Y., Liu, A.~H., Chang, S., Liao, Y.-L., Chuang, Y.-S.,
  Qian, K., Khurana, S., Cox, D., and Glass, J.
\newblock Parp: Prune, adjust and re-prune for self-supervised speech
  recognition.
\newblock In Ranzato, M., Beygelzimer, A., Dauphin, Y., Liang, P., and Vaughan,
  J.~W. (eds.), \emph{Advances in Neural Information Processing Systems},
  volume~34, pp.\  21256--21272. Curran Associates, Inc., 2021.

\bibitem[Li et~al.(2020{\natexlab{a}})Li, Wu, Su, and Wang]{li2020eagleeye}
Li, B., Wu, B., Su, J., and Wang, G.
\newblock Eagleeye: Fast sub-net evaluation for efficient neural network
  pruning.
\newblock In Vedaldi, A., Bischof, H., Brox, T., and Frahm, J.-M. (eds.),
  \emph{Proc. of the Computer Vision -- ECCV 2020}, pp.\  639--654, Cham,
  2020{\natexlab{a}}. Springer International Publishing.
\newblock ISBN 978-3-030-58536-5.

\bibitem[Li et~al.(2016)Li, Kadav, Durdanovic, Samet, and
  Graf]{Li2016handcraft}
Li, H., Kadav, A., Durdanovic, I., Samet, H., and Graf, H.~P.
\newblock Pruning filters for efficient convnets, 2016.

\bibitem[Li et~al.(2020{\natexlab{b}})Li, Gu, Mayer, Gool, and
  Timofte]{li2020group}
Li, Y., Gu, S., Mayer, C., Gool, L.~V., and Timofte, R.
\newblock Group sparsity: The hinge between filter pruning and decomposition
  for network compression.
\newblock In \emph{Proc. of the IEEE/CVF Conference on Computer Vision and
  Pattern Recognition}, pp.\  8018--8027, 2020{\natexlab{b}}.

\bibitem[Li et~al.(2020{\natexlab{c}})Li, Gu, Zhang, Van~Gool, and
  Timofte]{li2020dhp}
Li, Y., Gu, S., Zhang, K., Van~Gool, L., and Timofte, R.
\newblock {DHP}: Differentiable meta pruning via hypernetworks,
  2020{\natexlab{c}}.

\bibitem[Li et~al.(2021{\natexlab{a}})Li, Hao, Li, Xiong, and
  Chen]{li2021gnasr}
Li, Y., Hao, C., Li, P., Xiong, J., and Chen, D.
\newblock Generic neural architecture search via regression.
\newblock In Ranzato, M., Beygelzimer, A., Dauphin, Y., Liang, P., and Vaughan,
  J.~W. (eds.), \emph{Advances in Neural Information Processing Systems},
  volume~34, pp.\  20476--20490. Curran Associates, Inc., 2021{\natexlab{a}}.

\bibitem[Li et~al.(2021{\natexlab{b}})Li, Yuan, Niu, Zhao, Li, Cai, Shen, Zhan,
  Kong, Jin, et~al.]{li2021npas}
Li, Z., Yuan, G., Niu, W., Zhao, P., Li, Y., Cai, Y., Shen, X., Zhan, Z., Kong,
  Z., Jin, Q., et~al.
\newblock Npas: A compiler-aware framework of unified network pruning and
  architecture search for beyond real-time mobile acceleration.
\newblock In \emph{Proc. of the IEEE/CVF Conference on Computer Vision and
  Pattern Recognition (CVPR)}, pp.\  14255--14266, 2021{\natexlab{b}}.

\bibitem[Liben-Nowell \& Kleinberg(2007)Liben-Nowell and
  Kleinberg]{Nowell2007linkprediction}
Liben-Nowell, D. and Kleinberg, J.
\newblock The link-prediction problem for social networks.
\newblock \emph{Journal of the American Society for Information Science and
  Technology}, 58\penalty0 (7):\penalty0 1019--1031, 2007.

\bibitem[Liebenwein et~al.(2020)Liebenwein, Baykal, Lang, Feldman, and
  Rus]{liebenwein2020pfp}
Liebenwein, L., Baykal, C., Lang, H., Feldman, D., and Rus, D.
\newblock Provable filter pruning for efficient neural networks.
\newblock In \emph{International Conference on Learning Representations}, 2020.

\bibitem[Lillicrap et~al.(2016)Lillicrap, Hunt, Pritzel, Heess, Erez, Tassa,
  Silver, and Wierstra]{lillicrap2016ddpg}
Lillicrap, T.~P., Hunt, J.~J., Pritzel, A., Heess, N., Erez, T., Tassa, Y.,
  Silver, D., and Wierstra, D.
\newblock Continuous control with deep reinforcement learning.
\newblock In \emph{Proc. of the ICLR (Poster)}, 2016.

\bibitem[Lin et~al.(2017)Lin, Rao, Lu, and Zhou]{Lin2017RNP}
Lin, J., Rao, Y., Lu, J., and Zhou, J.
\newblock Runtime neural pruning.
\newblock In \emph{Proc. of the Advances in Neural Information Processing
  Systems}, pp.\  2181--2191, 2017.

\bibitem[Lin et~al.(2020)Lin, Ji, Wang, Zhang, Zhang, Tian, and
  Shao]{lin2020hrank}
Lin, M., Ji, R., Wang, Y., Zhang, Y., Zhang, B., Tian, Y., and Shao, L.
\newblock Hrank: Filter pruning using high-rank feature map.
\newblock In \emph{Proc. of the IEEE/CVF Conference on Computer Vision and
  Pattern Recognition}, pp.\  1529--1538, 2020.

\bibitem[Liu et~al.(2020)Liu, Ma, Xu, Wang, Tang, and Ye]{liu2020AutoCompress}
Liu, N., Ma, X., Xu, Z., Wang, Y., Tang, J., and Ye, J.
\newblock {AutoCompress}: An automatic dnn structured pruning framework for
  ultra-high compression rates.
\newblock In \emph{Proc. of the Artificial Intelligence Conference (AAAI)},
  pp.\  4876--4883, 2020.

\bibitem[Liu et~al.(2021)Liu, Liu, Lin, Dong, and Wang]{liu_learnable_2021}
Liu, Y., Liu, L., Lin, C., Dong, Z., and Wang, W.
\newblock Learnable {Motion} {Coherence} for {Correspondence} {Pruning}.
\newblock In \emph{Proc. of the {IEEE}/{CVF} {Conference} on {Computer}
  {Vision} and {Pattern} {Recognition} ({CVPR})}, pp.\  3237--3246, June 2021.

\bibitem[Liu et~al.(2019)Liu, Mu, Zhang, Guo, Yang, Cheng, and
  Sun]{liu2019metapruning}
Liu, Z., Mu, H., Zhang, X., Guo, Z., Yang, X., Cheng, K.-T., and Sun, J.
\newblock Metapruning: Meta learning for automatic neural network channel
  pruning.
\newblock In \emph{Proc. of the IEEE International Conference on Computer
  Vision}, pp.\  3296--3305, 2019.

\bibitem[Ma et~al.(2018)Ma, Zhang, Zheng, and Sun]{ma2018shufflenetv2}
Ma, N., Zhang, X., Zheng, H.-T., and Sun, J.
\newblock Shufflenet v2: Practical guidelines for efficient cnn architecture
  design.
\newblock In \emph{Proc. of the European conference on computer vision (ECCV)},
  pp.\  116--131, 2018.

\bibitem[Mehta et~al.(2020)Mehta, Hajishirzi, and Rastegari]{mehta2020dicenet}
Mehta, S., Hajishirzi, H., and Rastegari, M.
\newblock Dicenet: Dimension-wise convolutions for efficient networks.
\newblock \emph{IEEE Transactions on Pattern Analysis and Machine
  Intelligence}, 2020.

\bibitem[Ning et~al.(2020{\natexlab{a}})Ning, Zhao, Li, Lei, Wang, and
  Yang]{ning2020dsa}
Ning, X., Zhao, T., Li, W., Lei, P., Wang, Y., and Yang, H.
\newblock {DSA}: More efficient budgeted pruning via differentiable sparsity
  allocation.
\newblock In \emph{Proc. of 16th European Computer Vision Conference}, pp.\
  592--607. Springer, 2020{\natexlab{a}}.

\bibitem[Ning et~al.(2020{\natexlab{b}})Ning, Zheng, Zhao, Wang, and
  Yang]{ning2020generic}
Ning, X., Zheng, Y., Zhao, T., Wang, Y., and Yang, H.
\newblock A generic graph-based neural architecture encoding scheme for
  predictor-based nas.
\newblock In \emph{Proc. of European Conference on Computer Vision}, pp.\
  189--204. Springer, 2020{\natexlab{b}}.

\bibitem[Russakovsky et~al.(2015)Russakovsky, Deng, Su, Krause, Satheesh, Ma,
  Huang, Karpathy, Khosla, Bernstein, Berg, and Fei-Fei]{Olga2015ImageNet}
Russakovsky, O., Deng, J., Su, H., Krause, J., Satheesh, S., Ma, S., Huang, Z.,
  Karpathy, A., Khosla, A., Bernstein, M., Berg, A.~C., and Fei-Fei, L.
\newblock {ImageNet Large Scale Visual Recognition Challenge}.
\newblock \emph{International Journal of Computer Vision (IJCV)}, 115\penalty0
  (3):\penalty0 211--252, 2015.
\newblock \doi{10.1007/s11263-015-0816-y}.

\bibitem[Schlichtkrull et~al.(2018)Schlichtkrull, Kipf, Bloem, van den Berg,
  Titov, and Welling]{Schlichtkrull2018rgcn}
Schlichtkrull, M., Kipf, T.~N., Bloem, P., van den Berg, R., Titov, I., and
  Welling, M.
\newblock Modeling relational data with graph convolutional networks.
\newblock In Gangemi, A., Navigli, R., Vidal, M.-E., Hitzler, P., Troncy, R.,
  Hollink, L., Tordai, A., and Alam, M. (eds.), \emph{The Semantic Web}, pp.\
  593--607, Cham, 2018. Springer International Publishing.
\newblock ISBN 978-3-319-93417-4.

\bibitem[Schulman et~al.(2017)Schulman, Wolski, Dhariwal, Radford, and
  Klimov]{schulman2017ppo}
Schulman, J., Wolski, F., Dhariwal, P., Radford, A., and Klimov, O.
\newblock Proximal policy optimization algorithms, 2017.

\bibitem[Shi et~al.(2019)Shi, Pi, Xu, Li, Kwok, and Zhang]{Han2020NAS_oneshot}
Shi, H., Pi, R., Xu, H., Li, Z., Kwok, J.~T., and Zhang, T.
\newblock Bridging the gap between sample-based and one-shot neural
  architecture search with {BONAS}, 2019.

\bibitem[Tan \& Le(2019)Tan and Le]{tan2019efficientnet}
Tan, M. and Le, Q.
\newblock Efficientnet: Rethinking model scaling for convolutional neural
  networks.
\newblock In \emph{Proc. of the International Conference on Machine Learning},
  pp.\  6105--6114. PMLR, 2019.

\bibitem[Tang et~al.(2020)Tang, Wang, Xu, Tao, XU, Xu, and Xu]{tang2020scop}
Tang, Y., Wang, Y., Xu, Y., Tao, D., XU, C., Xu, C., and Xu, C.
\newblock Scop: Scientific control for reliable neural network pruning.
\newblock In Larochelle, H., Ranzato, M., Hadsell, R., Balcan, M.~F., and Lin,
  H. (eds.), \emph{Advances in Neural Information Processing Systems},
  volume~33, pp.\  10936--10947. Curran Associates, Inc., 2020.

\bibitem[Wang et~al.(2017)Wang, Zhang, Wang, and Hu]{wang2017SPP}
Wang, H., Zhang, Q., Wang, Y., and Hu, R.
\newblock Structured probabilistic pruning for deep convolutional neural
  network acceleration.
\newblock \emph{British Machine Vision Conference}, 2017.

\bibitem[Wang et~al.(2020)Wang, Wang, Cai, Lin, Liu, Wang, Lin, and
  Han]{wang2020apq}
Wang, T., Wang, K., Cai, H., Lin, J., Liu, Z., Wang, H., Lin, Y., and Han, S.
\newblock Apq: Joint search for network architecture, pruning and quantization
  policy.
\newblock In \emph{Proc. of IEEE/CVF Conference on Computer Vision and Pattern
  Recognition (CVPR)}, pp.\  2075--2084, 2020.

\bibitem[Wang et~al.(2021)Wang, Li, and Wang]{Wang2021cov}
Wang, Z., Li, C., and Wang, X.
\newblock Convolutional neural network pruning with structural redundancy
  reduction.
\newblock In \emph{Proc. of the IEEE/CVF Conference on Computer Vision and
  Pattern Recognition (CVPR)}, pp.\  14913--14922, June 2021.

\bibitem[Yang et~al.(2018)Yang, Howard, Chen, Zhang, Go, Sandler, Sze, and
  Adam]{yang2018netadapt}
Yang, T.-J., Howard, A., Chen, B., Zhang, X., Go, A., Sandler, M., Sze, V., and
  Adam, H.
\newblock Netadapt: Platform-aware neural network adaptation for mobile
  applications.
\newblock In \emph{The European Conference on Computer Vision (ECCV)},
  September 2018.

\bibitem[Yao et~al.(2021)Yao, Pi, Xu, Zhang, Li, and
  Zhang]{Yao2021JointDetNASUY}
Yao, L., Pi, R., Xu, H., Zhang, W., Li, Z., and Zhang, T.
\newblock Joint-detnas: Upgrade your detector with nas, pruning and dynamic
  distillation.
\newblock In \emph{Proc. of IEEE/CVF Conference on Computer Vision and Pattern
  Recognition (CVPR)}, pp.\  10170--10179, 2021.

\bibitem[Ye et~al.(2020)Ye, Gong, Nie, Zhou, Klivans, and
  Liu]{ye2020goodsubnet}
Ye, M., Gong, C., Nie, L., Zhou, D., Klivans, A., and Liu, Q.
\newblock Good subnetworks provably exist: Pruning via greedy forward
  selection.
\newblock In III, H.~D. and Singh, A. (eds.), \emph{Proc. of the 37th
  International Conference on Machine Learning}, volume 119 of \emph{Proc. of
  Machine Learning Research}, pp.\  10820--10830. PMLR, 13--18 Jul 2020.

\bibitem[Ying et~al.(2018)Ying, You, Morris, Ren, Hamilton, and
  Leskovec]{Ying2018DiffPool}
Ying, R., You, J., Morris, C., Ren, X., Hamilton, W.~L., and Leskovec, J.
\newblock Hierarchical graph representation learning with differentiable
  pooling.
\newblock In \emph{Proc. of the 32nd International Conference on Neural
  Information Processing Systems}, NIPS'18, pp.\  4805–4815, Red Hook, NY,
  USA, 2018. Curran Associates Inc.

\bibitem[Yu \& Huang(2019)Yu and Huang]{yu2019autoslim}
Yu, J. and Huang, T.
\newblock Autoslim: Towards one-shot architecture search for channel numbers,
  2019.

\bibitem[Yu et~al.(2021)Yu, Mazaheri, and Jannesari]{yu2020agmc}
Yu, S., Mazaheri, A., and Jannesari, A.
\newblock Auto graph encoder-decoder for neural network pruning.
\newblock In \emph{Proc. of the IEEE/CVF International Conference on Computer
  Vision (ICCV)}, pp.\  6362--6372, October 2021.

\bibitem[Zhang et~al.(2018{\natexlab{a}})Zhang, Ye, Zhang, Tang, Wen, Fardad,
  and Wang]{zhang2018unstructured}
Zhang, T., Ye, S., Zhang, K., Tang, J., Wen, W., Fardad, M., and Wang, Y.
\newblock A systematic {DNN} weight pruning framework using alternating
  direction method of multipliers.
\newblock \emph{ECCV}, 2018{\natexlab{a}}.

\bibitem[Zhang et~al.(2018{\natexlab{b}})Zhang, Zhou, Lin, and
  Sun]{zhang2018shufflenet}
Zhang, X., Zhou, X., Lin, M., and Sun, J.
\newblock Shufflenet: An extremely efficient convolutional neural network for
  mobile devices.
\newblock In \emph{Proc. of the IEEE conference on computer vision and pattern
  recognition}, pp.\  6848--6856, 2018{\natexlab{b}}.

\bibitem[Zhuang et~al.(2020)Zhuang, Zhang, Huang, Zeng, Shuang, and
  Li]{zhuang2020neuron}
Zhuang, T., Zhang, Z., Huang, Y., Zeng, X., Shuang, K., and Li, X.
\newblock Neuron-level structured pruning using polarization regularizer.
\newblock In Larochelle, H., Ranzato, M., Hadsell, R., Balcan, M.~F., and Lin,
  H. (eds.), \emph{Advances in Neural Information Processing Systems},
  volume~33, pp.\  9865--9877. Curran Associates, Inc., 2020.

\end{thebibliography}
\bibliographystyle{icml2022}

\newpage
\appendix
\onecolumn





\end{document}